\documentclass[letterpaper, 10 pt, conference]{ieeeconf}  

\IEEEoverridecommandlockouts                              
\usepackage{pgf}
\usepackage{tikz}
\usetikzlibrary{arrows,automata}
\usepackage{german}
\usepackage{listings}

\usepackage{graphicx} 
\usepackage{amsmath} 
\usepackage{amssymb}  

\title{\LARGE \bf
A Platform-independent Programming Environment for Robot Control
}

\author{ \parbox{6 in}{\centering Michael Reckhaus, Nico Hochgeschwender, Paul G. Ploeger and Gerhard K. Kraetzschmar
            Department of Computer Science\\
            Bonn-Rhine-Sieg University of Applied Sciences\\
            Sankt Augustin, Germany\\
            {\tt\small \{michael.reckhaus\}\{nico.hochgeschwender\}\{paul.ploeger\}\{gerhard.kraetzschmar\}@h-brs.de}}
}

\begin{document}

\maketitle
\thispagestyle{empty}
\pagestyle{empty}

\begin{abstract}
The development of robot control programs is a complex task. Many robots are different in their 
electrical and mechanical structure which is also reflected in the software. Specific robot software environments support 
the program development, but are mainly text-based and usually applied by experts in the field with profound knowledge of the target robot.
This paper presents a graphical programming environment which aims to ease the development of robot control programs. In contrast to existing graphical
robot programming environments, our approach focuses on the composition of parallel action sequences. The developed environment allows to schedule independent
robot actions on parallel execution lines and provides mechanism to avoid side-effects of parallel actions.
The developed environment is platform-independent and based on the model-driven paradigm. The feasibility of our approach is shown by the application of the sequencer to a simulated service robot and a robot for educational purpose.
\end{abstract}

\section{INTRODUCTION}
\label{sec:intro}
The development of robot control programs is a difficult, error-prone and complex exercise. 
At first, the developer needs a good understanding about the task which the robot should fulfill. 
This includes knowledge about the environment in which the robot is embedded and an idea
of the desired behavior of the robot. Second, platform-specific knowledge about the robot platform 
itself is needed. For instance, a vacuum cleaning robot is small and usually equipped with a cleaning device, whereas a service 
robot is tall and equipped with a manipulator to reach high cupboards and to grasp objects. This diversity in 
the mechanical and electrical structure reflects not only in the appearance and hardware design, but also in the software system. 
A robot application developer must be aware of this diversity in order to develop control programs in an optimal way. 
However, such a robot platform-oriented development of robot control programs leads (very often) to non-reusable 
control programs. Though, reuse of robot control programs is desirable, because the control program 
describes the logic of a specific task (e.g. an office delivery task) which could be applied 
on different robot platforms and reused (or partly reused) in similar tasks. To overcome the 
dependency of a specific platform (e.g. operating systems or even programming languages) the 
Software Engineering community introduced the concept of Platform-independent Models (PIM) 
and Platform-specific Models (PSM) within the model-driven software development paradigm. 
Here, models play a vital role to capture and abstract generic aspects from domain-specific aspects. 
Recently, the model-driven paradigm has been adopted by the robotics community. 
In~\cite{c3} Baer et al. introduced a platform independent modeling language providing modeling elements for
communication and collaboration infrastructures. The approach has been applied successfully for cooperative 
soccer playing robots. In Schlegel et al.~\cite{c4} a component-based robot software development approach has been applied to the model-driven paradigm in order to generate a wide variety
of target-specific source code (e.g. for embedded controllers) without modifying the platform-independent model.  
Another model-driven approach is described in~\cite{c7}. Here, Alonso et al. introduced a platform-independent 
component-model which captures certain aspects of a robotic software system in three distinct views, 
namely behavioral, structural, and component. The approach has been applied to generate 
Ada source-code for a Cartesian robot. 
\\[1.0ex]
However, to the best of our knowledge, the model-driven paradigm has been so far not applied to the development of 
platform-independent robot control programs, even so this domain could benefit from it. In particular, the development of 
robot control programs where concurrency is crucial (e.g. in behavior-based and hybrid control approaches as~\cite{c8}) is 
an error-prone task, because actions that may technically run in parallel can have side-effects. For instance, a manipulator can grasp an object while a camera processes images to detect objects. In such a situation the manipulator may block the field of view and the camera may not find the appropriate object. These pitfalls make the entry-level in robot programming very high and complicates the fast creation of robot control programs. Therefore, supporting tools and frameworks to ease the development are needed. 
\\[1.0ex]
One promising graphical programming environment with parallelism support is the Urbi Studio by the company Gostai~\cite{d14}. Urbi Studio provides a graphical state-machine based programming environment which is bound to their custom programming
language urbiscript. In addition to state-machine based programming Urbi Studio provides a time-based graphical sequencer to initialize and set variables over time -- a specialty provided by the framework. 
However, the environment is tightly bound to the Urbi execution engine which allows to monitor the program execution in the graphical environment. The company aims to make all robots compatible, which leads into a concrete implementation and C++ wrapping of the robot framework to the Urbi framework. Another graphical programming environment which also  provides the possibility to describe parallel sequences is the Nao Choregraphe from Aldebaran Robotics~\cite{d15}. It is delivered with their Nao robot. This environment allows flow diagram based programming of Nao robots and provides a time-based sequencer to create motor control commands. The environment is able to execute code on the Nao robot either in Urbi or the Python language.\\[1.0ex]
In this paper we present a programming environment for graphical sequencing of robot actions. 
The approach is based on the model-driven paradigm. Exchangeable domain-specific languages 
are used to generate platform-specific source code for a wide variety of robots without changing the
control logic. The paper is structured as follows. In Section~\ref{sec:problemstatement} the robot control 
approach, namely the sequencing of concurrent actions is introduced. Our model-driven approach to achieve 
platform-independence is described in Section~\ref{sec:approach}. The application of the graphical sequencer is 
shown in Section~\ref{sec:usecases} by the application to a simulated service robot and a a robot for educational purpose. In Section~\ref{sec:conclusion}
our approach will be discussed and we conclude with some lessons learned.

\section{ACTION SEQUENCING FOR ROBOT CONTROL}
\label{sec:problemstatement}
We are interested in the programming of robot tasks. For instance, the autonomous delivery of documents in an office environment. 
A task $\mathbf{T} = \{\mathbf{a_{1}}, \mathbf{...}, \mathbf{a_{n}}\}$ is composed of a set of finite actions (e.g. grasping an object, 
following a person, or moving to a location) from the set of actions $\mathbf{A}$ and $\mathbf{T} \subseteq \mathbf{A}$. 
Each action $\mathbf{a} \in \mathbf{A}$ is a tuple of the following form
\begin{equation*}
	\mathbf{a} = \langle \mathbf{n}, \mathbf{C} \rangle
\end{equation*}
where $\mathbf{n}$ is an unique name of the action and $\mathbf{C}$ is a set of execution constraints $\mathbf{e}$. Furthermore, a global data space $\mathbf{D}$ is 
used to store global variables which are needed, modified and shared by the actions. An execution constraint is formally defined as 
\begin{equation*}
	\mathbf{e} = \langle \mathbf{\Lambda}, \mathbf{a} \rangle \quad \mathbf{a} \in \mathbf{A} 
\end{equation*}

where $\mathbf{\Lambda}$ is an execution operator which constraints the execution of the action $\mathbf{a}$ in a temporal manner. 
In our case $\mathbf{\Lambda}$ describes the order of actions. Hence, our robot control approach may be described as a dependency 
graph \cite{d16} with nodes as actions and edges as execution constraints describing the predecessor and successor relationship between actions. 
Figure~1 shows an example of such an execution order. Here, the action \textbf{A}, \textbf{B}, and \textbf{C} are executed initially, since no constraints are
defined. Action \textbf{D} is constrained. \textbf{D} must only start as soon as the predecessors \textbf{A} and \textbf{B} finished. Furthermore, the action \textbf{E} is constrained. 
\textbf{E} is the last action and must only be executed, if all predecessors (namely \textbf{A}, \textbf{C} and \textbf{D}) finished. 

\begin{figure}
\begin{center}
\begin{tikzpicture}[->,>=stealth',shorten >=2pt,auto,node distance=2cm,
                    semithick]
  \tikzstyle{every state}=[fill=blue,draw=none,text=white]

  \node[state] (C) {$\mathbf{C}$};
  \node[state] (B) {$\mathbf{B}$};
  \node[state] (C) [below of=B] {$\mathbf{C}$};
  \node[state] (B) [above of=C] {$\mathbf{B}$};
  \node[state] (A) [above of=B] {$\mathbf{A}$};
  \node[state] (D) [right of=B] {$\mathbf{D}$};
  \node[state] (E) [right of=D] {$\mathbf{E}$};

  \path (A) edge node {} (E)
  	   (A) edge node {} (D)
  	   (B) edge node {} (D)
	   (C) edge node {} (E)
	   (D) edge node {} (E);
  \end{tikzpicture}
\caption{An dependency graph representation of the sequence of robot actions}
\end{center}
\label{fig:graph}
\end{figure}
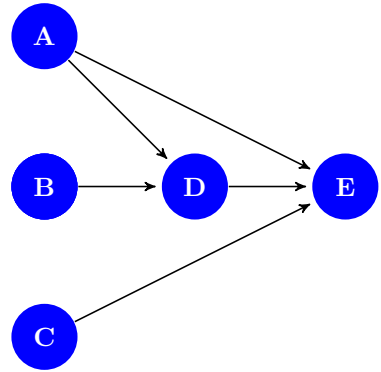

\section{APPROACH}
\label{sec:approach}
In this section we will describe our approach that is based on a model driven paradigm. We show how the concepts of meta-modeling and domain-specific languages are applied to develop a platform and robot system independent programming environment which focuses on parallel action sequences.

\subsection{Requirements}
The programming environment shall assist robot programmers to develop robot control sequences as those described in Section~\ref{sec:problemstatement}. Thereby it shall fulfill two main requirements. First, a simplification of action sequences, so that no experts in the field are required. In order to achieve this, an intuitive graphical program editor must be provided. Besides the simplification of action sequence creation the environment must assist the developer to reduce failures that occur due to parallelism in such sequences, which could be e.g. a manipulator that grasps an object and blocks the field of view of a camera recording images.

Secondly, it is required that the environment is independent of the target robot. It must be possible to describe robot action sequences for any robot platform. Thereby, different robot classes can have different programming elements e.g. a vacuum cleaning robot may not have a manipulator and a programmer should not have the possibility to use a grasp action on these robot types. The programming elements must be adaptable and even the same control programs shall be applicable to different robots.

\subsection{Meta-model}
\begin{figure}[htbp]
		\begin{center}
		\includegraphics[width=0.5 \textwidth]{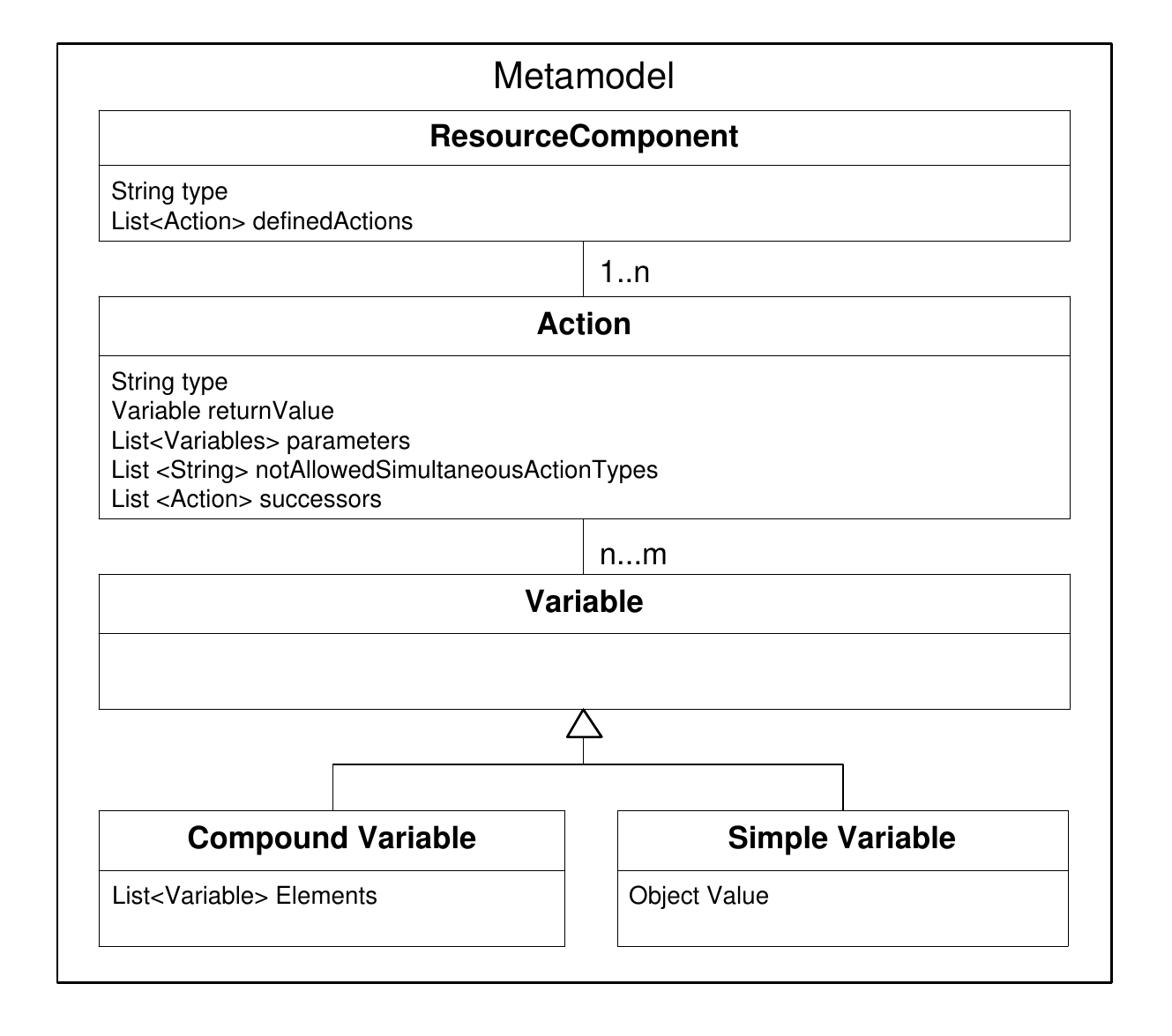}
		\end{center}
		\caption{Sketch of the meta-model}
		\label{fig:PaperMetaModel}
\end{figure}

The programming model is structured by a meta-model which describes the structure of the introduced action sequences, namely the abstract syntax and semantics. The concept follows the scheme of Model-Driven Software Development (MDSD) defined by Stahl and V\"olte~\cite{c9}. The meta-model describes the basic programming elements `Action', `Resource Component', and `Variable'. See Figure~\ref{fig:PaperMetaModel}. 
\begin{itemize}
	\item \textbf{Action:} An action is a certain thing that a robot can do, like `move to a position', `set motor speeds' or `capture an image'. Actions can work on a set of predefined global variables which are used as parameter or return values, comparable to 		
	function calls. Actions can be put into an execution sequence as defined in Section~\ref{sec:problemstatement}. Further, to assist concurrent programming, every action has a list of concrete actions, which are not allowed to be scheduled simultaneously. 
	Every action belongs to a resource component.
	\item \textbf{Resource Component:} A resource component enables the usage of a certain action set. For instance, the definition of a concrete resource component `manipulator' enables the programmer to use the actions `move arm to pose' and `grasp'. If a 
	robot does not have a manipulator, it also should not enable the programmer to use those actions. A resource component can schedule a single action at a time in a serial manner. Only the use of multiple resource components allows the creation of 
	parallel action sequences. So, if a robot has only a single resource component `manipulator' it shall not be able to execute multiple `grasp' actions simultaneously. An additional resource component like human machine interface would allow simultaneous 
	speech output to grasping actions. The term resource component refers not necessarily to a specific hardware device. It may also represent a computational unit like a functional library for planning, numerical computation, mapping or navigation. 
	\item \textbf{Variable:} The meta-model further contains the structure for variables that are globally defined. Variables are either simple or arbitrarily cascaded like structs in the C language.  
\end{itemize}

\subsection{Domain-specific language}
\begin{figure}[htbp]
		\begin{center}
		\includegraphics[width=0.5 \textwidth]{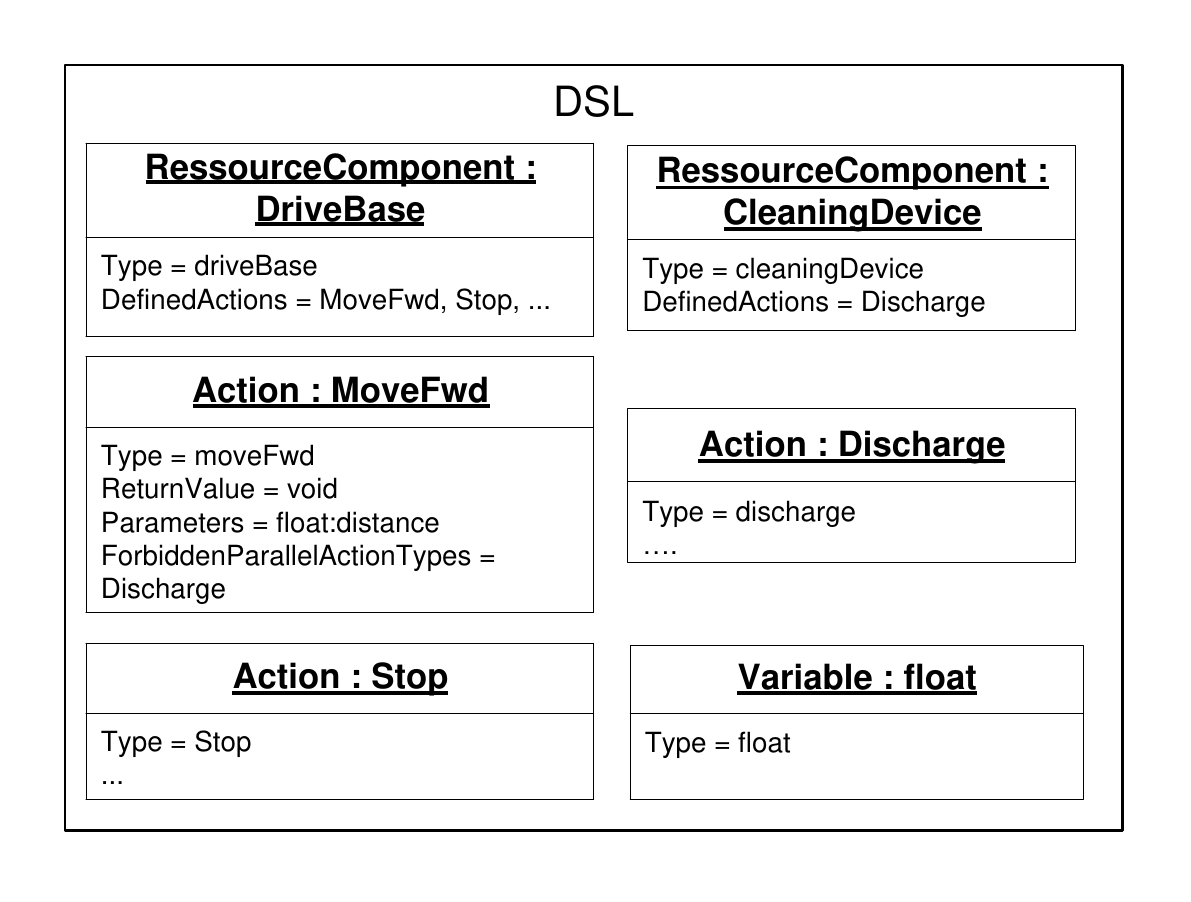}
		\end{center}
		\caption{An example excerpt of a concrete domain-specific language for a vacuum cleaning robot}
		\label{fig:PaperExampleDSL}
\end{figure}

The meta-model defines abstract constructs and does not define concrete functionality. Only the abstract elements action, resource component and variable are described. The concrete instances which are available to the programmer are formalized by the Domain Specific Language (DSL). The DSL can be described for different types of robot classes. Every DSL has to define which concrete resource components are provided by a specific robot class e.g. vacuum cleaners. For every resource component it must be defined which specific types of actions are provided by the resource. Those actions must each explicitly declare which other actions are not allowed to run simultaneously. Further, the DSL specifies the possible variables that a programmer can use. As mentioned, those can be basic variables like integer or float, or more complex structures build from other defined variables like the structure of an address. An example of a concrete DSL is sketched in Figure~\ref{fig:PaperExampleDSL}. It shows an excerpt of a DSL for a vacuum cleaning robot and defines the resource components \emph{DriveBase} with the actions \emph{MoveFwd} and \emph{Stop} and the resource component \emph{CleaningDevice} with the action \emph{Discharge}. The shown DSL has adds a parallelism constraint. It describes that the action \emph{moveFwd} may never run simultaneously to the action \emph{discharge}. This removes the semantical error that a vacuum cleaning robot could move while distributing all collected dirt on the floor.  

Finally, to get running robot code, a template-based code generator is used. The code generator is the first place in the process where a concrete robot systems is targeted since the program sequence and DSL can be defined for a complete robot class. Every resource component and every action can have an arbitrary number of code templates. Those templates must be written based on the robot framework and programming language. They must implement the action sequences for the robot. This makes it possible to generate code for different robot systems from the same sequence. Further, an arbitrary number of main templates can be specified. This may be necessary for robot code in e.g. C++ where multiple code files are required (source files and headers). Further, it enables the definition of code templates in different languages or frameworks for the same robot types but different concrete instances.

\begin{figure}[htbp]
		\begin{center}
		\includegraphics[width=0.5 \textwidth]{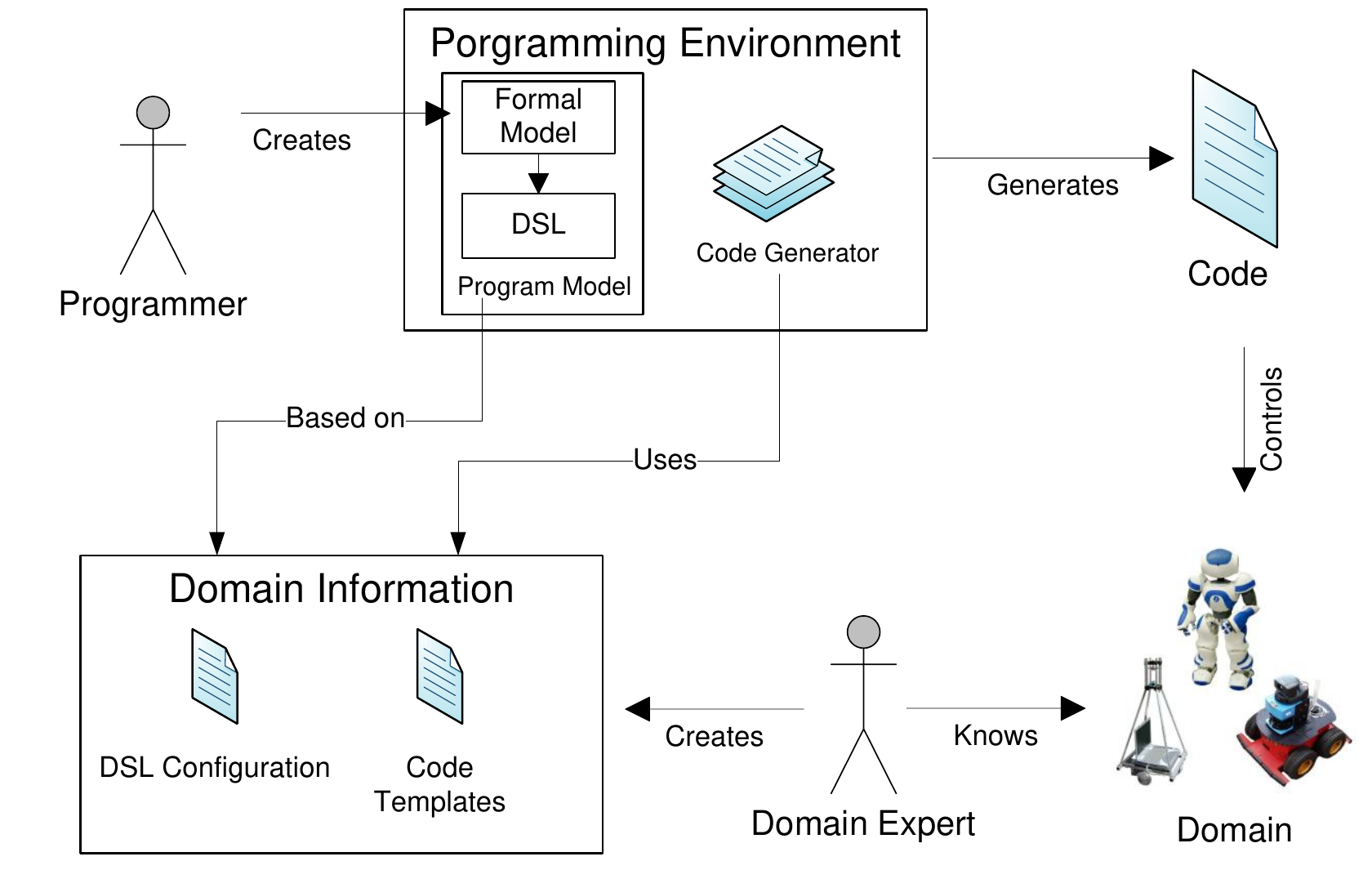}
		\end{center}
		\caption{System Concept in MDSD terminology}
		\label{fig:SystemConceptMDSD}	
\end{figure}

An overview of the system in the MDSD terminology is shown in Figure~\ref{fig:SystemConceptMDSD}. The figure shows the relation between the action sequence programmer, the programming environment and configuration as well as the domain expert that describes the domain specific elements. 
\begin{itemize}
	\item A \textbf{programmer} can create a concrete program model respective formal model with the programming environment which is an action sequence for a robot class. It is based on a DSL that contains the description of the concrete programming 
	elements for the robot class. This DSL is based on a meta-model which provides structure and is the same for any formal model and robot.
	\item The \textbf{domain expert} specifies the DSL via a configuration file. Profound knowledge in programming a certain robot respective robot class is needed. Moreover, the domain expert has to provide the code templates for a specific robot to enable the translation from formal model to specific robot code.

\end{itemize}

\subsection{Implementation}
The system is developed with Java and the Eclipse Rich Client Platform RCP~\cite{c13} and is operating system independent. The graphical editor is implemented with the Eclipse Graphical Editing Framework~\cite{c11} and allows the creation of the described action sequences. The meta-model is formalized as a set of Java classes while the DSL configuration can be separately loaded as XML files. Moreover, the existing sequences can be robot independent serialized to XML. The editor validates the generated action sequences on completeness, e.g. are all names unique, variables instantiated and parameters set and verifies if the defined parallelism constraints are not violated.
For code generation the Apache Velocity Engine~\cite{c12} is integrated. This code genaration library was chosen because of its very clear and powerful template language. The amount, location and names of the templates are also configurable with XML. 

\section{USE CASES}
\label{sec:usecases}
\subsection{Service robotics use case}
\begin{figure}[htbp]
	\centering
		\includegraphics[width=0.5 \textwidth]{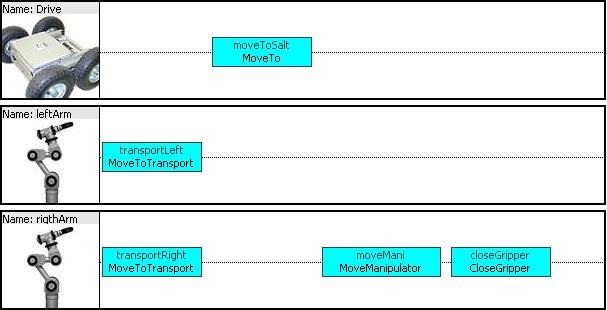}
		\caption{Sample sequence of a service robot grasping process}
		\label{fig:sampleMRDS}
\end{figure}

In the first use case a simulated service robot system with a holonomic drive system and two manipulators has been integrated in the environment. An example sequence which makes the robot move to a certain position and grasp an object is shown in Figure~\ref{fig:sampleMRDS}.  For this, a minimalistic DSL containing the resource components for each gripper and the drive base has been designed with actions for closing a gripper, moving the arm to certain positions and moving the robots base. An excerpt of the DSL is given in Listing~\ref{list:dslconfigscheduler}.

\begin{lstlisting}[stepnumber = 0, caption= DSL configuration of a MoveManipulator action,language=XML, basicstyle=\ttfamily\fontsize{8}{12}\selectfont,label = list:dslconfigscheduler]{}  
<ResourceComponent type="Manipulator">
 <Action returnType="String"
  actionIdentifier="MoveManipulator">
  <ParameterList>
   <Parameter type="Vector3" name="targetPose">
   </Parameter>
   <Parameter type="Vector3"name="orientation">
   </Parameter>
  </ParameterList>
  <NotAllowedSimultaneousActionTypes>
   <NotAllowedSimultaneousAction type="MoveTo">
   </NotAllowedSimultaneousAction>
  </NotAllowedSimultaneousActionTypes>
 </Action>
</ResourceComponent>
\end{lstlisting}

To generate code, every action and resource component has its custom code template. An excerpt of the moveManipulator action is given in Listing~\ref{list:MRDSTemplateMoveManipulator}. The template uses keywords of the velocity engine which are indicated by a '\#' for control structures and '\$' for data access from the formal model. Listing~\ref{list:MRDSGeneratedMoveManipulator} shows the generated code that results for this template.

\begin{lstlisting}[stepnumber = 0, caption= C\# code template for a MoveManipulator action, tabsize=2,  basicstyle=\ttfamily\fontsize{8}{12}\selectfont, label = list:MRDSTemplateMoveManipulator]{}
//Create list of parameters
parameters = new List<ParameterVariable>();
//fill list of parameters 
#foreach($Parameter in  $Action.getParameters())
//Add previous initialized variables
parameters.Add(getVariable(
  "$Parameter.getVariable().getName()"));
#end
//Create robot specific action
ExecutionElement $Action.getName() = 
	new ExecElement(MOVE_MANIPULATOR, parameters));
\end{lstlisting}

\begin{lstlisting}[stepnumber = 0, caption= Generated code for a manipulator moveTo action in C\#, tabsize=2,  basicstyle=\ttfamily\fontsize{8}{12}\selectfont, label = list:MRDSGeneratedMoveManipulator]{}
//Create list of parameters
parameters = new List<ParameterVariable>();
//Add previous initialized variables
parameters.Add(getVariable("targetPose"));
parameters.Add(getVariable("orientation"));
//Create robot system specific action
ExecutionElement MoveMani = 
	new ExecElement(MOVE_MANIPULATOR, parameters));
\end{lstlisting}

\subsection{Educational use case}
\begin{figure}[htbp]
	\centering
		\includegraphics[width=0.5 \textwidth]{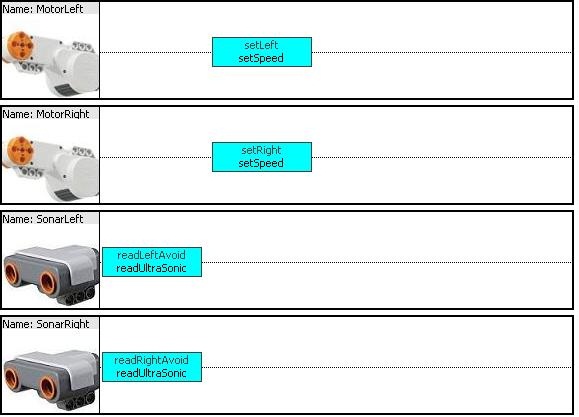}
		\caption{Sample sequence of an obstacle avoiding Lego NXT robot}
		\label{fig:sampleLegoObstacle}
\end{figure}

A further use case based on a different configured DSL has been implemented with the same programming environment. This DSL configures the environment to program a Lego Mindstorm NXT educational robot. The robot was constructed with a left and right ultra sonic sensor, that are oriented at the robot front. The robot is actuated by a differential drive base. This use case shows the implementation of a Braitenbergs vehicles~\cite{c10} so that the robot avoids obstacles. The program sequence is shown in Figure~\ref{fig:sampleLegoObstacle}. In Braitenberg vehicles the sensor readings are directly mapped to actuators which can result in complex reactive behaviors. For the use case the robots left sonar is mapped to the left motor and the right sonar is mapped to the right motor. The closer the robot senses an object on its right sensor, the higher the sensor value respective right motor speed which results in a turning behavior away from the obstacle. In the sequence shown in the figure, the sensors are read simultaneously and their results are passed to the left and right motors. This is done by two global variables storing the left and right ultra sonic sensor readings. After reading, these variables are used by the motors to set the speed. The left sensor shares a variable with the right motor and the right sensor shares a variable with the left motor. If an obstacle is detected in the left sensor, the motor on the right gets slower since it acts directly on the sensor reading that gets smaller by closer obstacle distance. This results in a turn away from the obstacle. The sequence is translated to NXC code that must be compiled and transfered to the robot.

\section{CONCLUSION}
\label{sec:conclusion}This paper presented a programming environment that allows the graphical programming of robot action sequences comparable to dependency graphs. Main focus of the environment is the simplification of robot programming especially in the sense of parallel program development. The designed system is based on the model-driven software development aspects which makes it possible to re-use it with any robot system or even transform the same described program into code for different robot platforms. In order to achieve this, a meta-model has been designed while robot specific DSLs and code generation settings can be loaded via XML configuration files. Two exemplary use-cases were described in which a a service robot and a robot for educational purpose were programmed with the presented environment.
The environment is successfully used in lectures to teach students the programming of Lego NXT Robots. There it has been proven as a valuable tool to teach parallel programming on real robots.

In the future we will develop domain-specific languages and code generation templates for different robot platforms so for example our service robot Johnny Jackanapes~\cite{c14} which is participating in the service robot competition RoboCup@Home. Further, the programming model will be enhanced by decisional elements to allow a state machine oriented programming.

\section{ACKNOWLEDGMENTS}
The research leading to these results has received funding from the European Community's Seventh Framework 
Programme (FP7/2007-2013) under grant agreement no. FP7-ICT-231940-BRICS (Best Practice in Robotics). 


\end{document}